\documentclass[conference,compsoc]{IEEEtran}
\IEEEoverridecommandlockouts
\ifCLASSOPTIONcompsoc
  \usepackage[nocompress]{cite}
\else
  \usepackage{cite}
\fi
\usepackage[pdftex]{graphicx}
\graphicspath{{./}}
\DeclareGraphicsExtensions{.pdf,.jpeg,.pdf}

\ifCLASSINFOpdf
\else
\fi
\usepackage{amsmath}
\begin{document}
%
\title{Swarm Differential Privacy for Purpose Driven Data-Information-Knowledge-Wisdom Architecture}

\makeatletter
\newcommand{\linebreakand}{%
  \end{@IEEEauthorhalign}
  \hfill\mbox{}\par
  \mbox{}\hfill\begin{@IEEEauthorhalign}
}
\makeatother

\author{\IEEEauthorblockN{Yingbo Li}
\IEEEauthorblockA{
Hainan University\\
lantuzi@aliyun.com}

\and
\IEEEauthorblockN{Yucong Duan* \thanks{*Corresponding author: duanyucong@hotmail.com}}
\IEEEauthorblockA{
Hainan University\\
duanyucong@hotmail.com}

\and
\IEEEauthorblockN{Zakaria Maamar}
\IEEEauthorblockA{
Zayed University\\
Zakaria.Maamar@zu.ac.ae}

\linebreakand

\and
\IEEEauthorblockN{Haoyang Che}
\IEEEauthorblockA{
Zeekr Group\\
chehy@hotmail.com}

\and
\IEEEauthorblockN{Anamaria-Beatrice Spulber}
\IEEEauthorblockA{
Visionogy\\
anne@visionogy.com}

\and
\IEEEauthorblockN{Stelios Fuentes}
\IEEEauthorblockA{
Leicester University\\
stelios.fuentes@gmx.co.uk }

}


%


\maketitle

\begin{abstract}Privacy protection has recently been in the spotlight of attention to both academia and industry. Society protects individual data privacy through complex legal frameworks. The increasing number of applications of data science and artificial intelligence has resulted in a higher demand for the ubiquitous application of the data. The privacy protection of the broad Data-Information-Knowledge-Wisdom (DIKW) landscape, the next generation of information organization, has taken a secondary role. In this paper, we will explore DIKW architecture through the applications of the popular swarm intelligence and differential privacy. As differential privacy proved to be an effective data privacy approach, we will look at it from a DIKW domain perspective. Swarm Intelligence can effectively optimize and reduce the number of items in DIKW used in differential privacy, thus accelerating both the effectiveness and the efficiency of differential privacy for crossing multiple modals of conceptual DIKW. The proposed approach is demonstrated through the application of personalized data that is based on the open-sourse IRIS dataset. This experiment demonstrates the efficiency of Swarm Intelligence in reducing computing complexity.
\end{abstract}

\begin{IEEEkeywords}
Swarm Intelligence, Differential Privacy, DIKW, Particle Swarm Optimization, Privacy Protection
\end{IEEEkeywords}

%
\IEEEpeerreviewmaketitle

\section{Introduction}

Artificial intelligence \cite{ai}\cite{pml} has been in limelight as a result of the abundant data stemmed from big data and acquired from multiple industries: healthcare, lifelog, Internet and Internet of Things (IoT). From a hardware perspective, each organization used to build their own infrastrcuture to store the daily increasing data. However, this proves to be difficult for many non-IT organizations that lack the necessary skills, resources and budgets. The emergence of cloud service providers \cite{cloud} such as AWS, Google, or Microsoft Azure brought a solution to the complex server and hardware management issues that common public experience and it even solved some of the biggest software concerns. Thus, Big Data became a popular application in softwares. However, it also brings new challenges with it: how can we better use the large volume of data that we stored? It is also one of many reasons for why artificiall intelligence became a popular science discipline: artificial intelligence can now have enough training data and applications demand. 

In addition to the data analysis such as feature extraction and prediction, information processing based on big data also demands better organization of data. The knowledge graph \cite{kg} \cite{graph}  \cite{tgraph} can successfully be used for semantic data organization. It organizes the knowledge from two aspects: completeness and correctness. Knowledge graph has originally been proposed by Google for Web Semantics and Web search. It organizes the data well however, a more complex data configuration is needed as not every data should be put in the same category or modal. For example, the answer to the question - why is the earth round is more related to the knowledge modal, but not to the single data type "earth" or "round". Therefore, in recent years, researchers proposed the Data-Information-Knowledge-Wisdom (DIKW) \cite{dikw}\cite{Ackoff} \cite{dikv}\cite{dikw01}\cite{dikw02} architecture to separate the collected data into multiple modals of DIKW. A data modal and information modal contain specific data and information such as weather temperature, time, web service\cite{dikw03}, biological resource\cite{dikwa1} and location\cite{dikwa2}, whilst other DIKW modals contain further semantics and abstractive characteristics.

Data privacy \cite{privacy}\cite{bdp2}  is becoming a popular subject for the common society and extensive research is being carried out on data privacy legislation in The European Union and The United States \cite{gdpr}\cite{ccpa}. The large volume of data within Big Data demands better privacy protection \cite{bdp} \cite{ds} \cite{dsp}. Regulations require people to carefully consider the use of indiviual related data, or the data that could be used to identify the specific individual user or the specific group of users. Data masking and data encryption are commonly used techniques for data privacy. However, maximizing the protection of individual data, goes against the economical goal of maximizing the technical power of individual data usages such as personal recommendations. Additionally, it brings challenges to the social goal of maximizing the social welfare through the use of data collections from individuals - group data usage such as pandemic diseases, treatment discovery, etc. To optimize the effectiveness of privacy protection, an universal approach to uniformly merge and balance both the purposes in technical data protection methods and the purposes in expanding information usages embedded in restricted data and knowledge, is necessary. This requisite is justified in the context of already existent multiple data modal expressions of the same piece of information \cite{dikwprivacy}. This data modals can take the form of individual's health condition information that can be recorded by numerious data indicators, whilst some data can be derived from knowledge formulas. A possible advantage brought by the interchangeability of the DIKW modals is the improved efficiency in data usage space through modals addition as alternative options to the original data entities. For example it can use alternative data to replace or simulate the missing data, or deriving data through calculation from formulas instead of exploring the vast data space,etc. The privacy protection of DIKW has not attracted much attention in the research domain until recently and as such DIKW is a developing discipline. 

In this paper we propose a novel framework for DIKW privacy protection by joining the optimization algorithm - swarm intelligence and the data privacy algorithm - differential privacy. We start by reviewing the state of the art for big data, DIKW and  data privacy. Subsequently, we will introduce the differential privacy concept with the view towards its applications in the privacy protection of DIKW. And at last, we will use Swarm Intelligence concept as a method to reduce the time complexity in data privacy of DIKW. Finally we will conduct the case study and the corresponding experiment for the proposed framework. 

\section{State of the art from big data to DIKW}

The volume of the data that is available nowadays is increasing dramatically, day by day, with industries such as smart cities, health or Internet amongst many other examples. The increasing amount of data that is currently available has been the main reason behind the increasing popularity of disciplines such a Big Data, Artificial Intelligence or Cloud Computing. Since big data requires  much higher storage capablities for large volume of data and higher data security, which exceeds the capablities of most organizations and companies, cloud infrastructure providers such as AWS, are becoming more and more popular in the current technology world. In order to satisfy the required storage and computing, especially the real-time computing and parallel computing to the big data, different open-source softwares of big data were developed such as Hadoop, HBase, Yarn etc. \cite{hbase}. 

Big data is widely applied in multiple industries\cite{Oussous}: Internet of Things (IoT) \cite{gao1}\cite{gao2}\cite{gao3}, Edge computing \cite{yin} \cite{yang}, health, smart city, transportation, or social monitoring. These industries generate daily large volume of data. In IoT wireless sensors, NFC \& GPS generate data everywhere and anytime. Health industry, with plenty of classical samples for health problems such as the disease, the sleepness, and other similar data samples, are typical applications of Artificial Intelligence (AI) because it can provide enough training data to the researcher. Smart City \cite{ppd} with multpile types of sensors for electricity, water or security applications could provide abudant information to the city that can be used to optimize the industry and the consumer usage. Autonomous vehicles for example have sensors all over the vehicle that highly depend on the storage and the real-time computing of big data in order to predict and decide the real-time behaviors of vehicles that are required in driving, road security and other similar examples. The social monitoring for both private purpose and public purpose from social media are becoming increasingly important industries that attract the interest of researchers. Private institutions lean on mining the data from social media to make product decisions, while the government uses social media as a data source for prediction and optimization of opinion polls.

Like many other disciplines, big data faces many challenges as it rapidly develops \cite{Oussous}. First, although big data means the large volume of the data, the data from specific datasets for specific problems in specific organizations is not enough. A second problem in big data is the data collection. For example, the weather data in one mountain needs to set up enough number of weather sensors to collect data. This cannot be conducted easily, even though there are open-source weather datasets. Consequently, the data openness and sharing is another obstacle that exists in the current commercial and research world of big data. For example, the sale of data by the bestsellers on Amazon is only known by Amazon; for other organizations it is diffcuilt to obtain this private datasets. In addition, even with enough data in specific datasets there are factors that can still cause problems of imbalance. The data imbalance is a common problem in the domain of big data \cite{bdp}, especially in the computation of data by machine learning, because the imbalanced data will cause the wrong prediction and classification in deep learing \cite{imbalance}. In order to resolve the data imbalance, we need technologies to clean and rebalance the uneven distribution of the data \cite{Ashrapov}. A popular approach to rebalance the data is the sampling of the dataset to create a balanced sub-dataset.

Big data is fundamental for machine learning and especially deep learning \cite{bdp}. Machine learning is often used to discover the hidden and semantic knowledge. Deep learning in machine learning  has been proven to be successful in multiple domains \cite{zhang}, such as Natural Language Processing, speech recognition, image classification, image recognition, and the lists can continue. Typical deep learning models include stacked, auto-encoder, convolutional neural network and recurrent neural network. Recently, transfer learning \cite{transfer} became popular because it extends the applications of deep learning and resolves deep learning problems with limited datasets by transfering the trained models to other problems.

Big data brings multiple opportunities to research and industry \cite{security}. However, there are still many issues of data security and privacy risk \cite{piq}. The data security consists in the security of the source of the data. Since each item of big data is from one individual sensor or person, it is possible that the data is obtained without the corresponding permission that breaks the law in many countries. Even if the data collecton is legal, it still brings the risk of data leakage, which often happened for many credit card and telephone numbers accounts. Considering the above aspect of data security, it is normal to be concerned with the data privacy. Since the data is often from and related to a specific person, laws like EU GDPR \cite{gdpr} begin to limit the usage of the data with personal privacy laws. The deep learning technologies using big data can cause many problems. Face recognition in the video survillance \cite{face}, which is commonly used in different security applications, causes the concern of personal physical tracking. Another technology, Deepfake \cite{deepfake}, that can facilitate the image and video falsification, caused unimagined problems in the past especially to many known personalities and celebrities. In order to resolve the worry from common public for data security and privacy, researchers are working on both the legislation of data privacy law and the technical approach. Data masking \cite{masking}  is one common technical approach to separate the data from a specific person and  specific group of people. The data awareness requires the common general public to know the importance of their personal data and that it is being conducted by the researcher and the media. Many countries added data awareness laws for data privacy such as the recent EU GDPR, and set up administrative agencies that are concerned with data privacy.

With the increase of the data volume, the data of each index from multimedia and multi-modal information becomes more complex and semantically meaningful. Consequently, the cross-relation among different indicies of information brings the semantic information in the $Data$, $Information$, $Knowledge$ and $Wisdom$ (DIKW) architecture\cite{dikw} \cite{dikw2}\cite{dikw4}. DIKW has been a successful framework to combine and elevate the multimodal data into the models of $Information$, and $Knowledge$. With the development of the society and technology, the information has been the widest notion in DIKW. The data refers to the specific data stored in the database. The knowledge could be infered from data and information, while the wisdom covers the notions of the data and information as part of the knowledge. DIKW could represent 5 Ws: $Who, What, When, Where$, and $Why$. Data linguistically represents true or false statements. $Information$ is the polysemantic concept such as the entropy, signal information, and semantic information. $Knowledge$ is the collection of "know-that's" like this: New York city is located in New York State, USA. Wisdom is a philosopher notion but $Wisdom$ in DIKW is known how to control systems. In data model we can know who, when and where, While in information and knowledge models we could probably infer what and how \cite{graphdik} by knowledge graph \cite{kg} \cite{gdesign}. The term "knowledge graph" was initially proposed by Google in 2012 and it used on semantic web \cite{kgsw2}\cite{sw3}. Until now many succesful knowledge graphs \cite{gbtia} such as DBpedia, YAGO, and Freebase have been proposed and used in real applications. Knowledge graph is composed of entities - the nodes of the graph, and the relations - the edges of the graph. One example of knowledge graph \cite{Existence} is shown in Fig. \ref{fig:kg}, and it is easy to conclude that it is useful in the inference, the recommendation and the search engine. Recently knowledge graph \cite{skg}\cite{skg2}\cite{kgr} is popularly used in  deep learning by converting the entities and relations of knowledge graph to a continuous vector space, named knowlege graph embedding \cite{kge}. In addition, knowledge graph was introduced to the domain of DIKW structure by $DataGraph$, $InforamtionGraph$ and $KnowlegeGraph$ \cite{graphdik}.

\begin{figure}[ht]
\centering
\includegraphics[width=0.7\columnwidth]{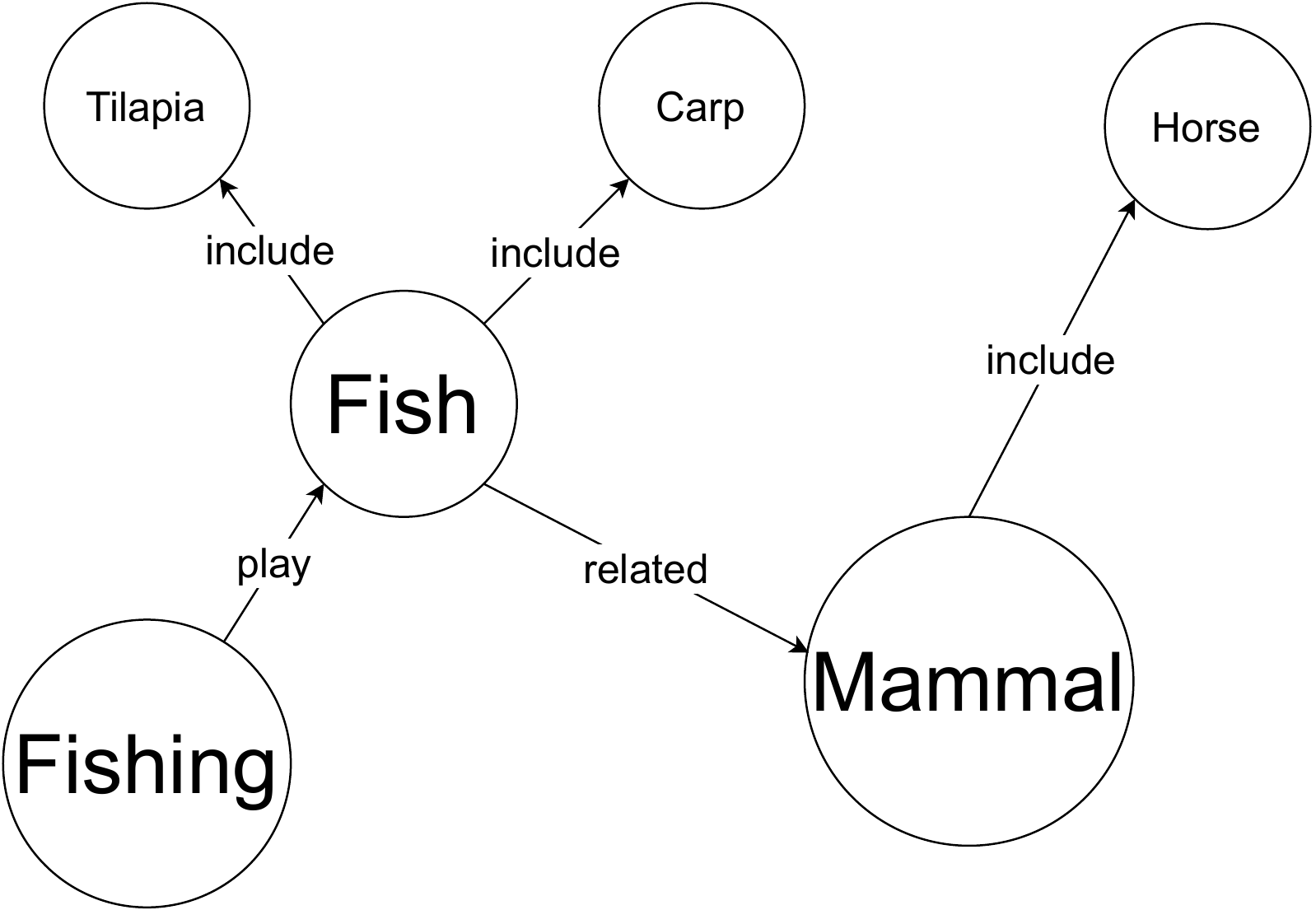}
\caption{Knowledge Graph}
\label{fig:kg}
\end{figure}

Since DIKW structure is useful to the inference, the leakage of partial data, information and knowledge in DIKW could lead to the inference of more information in the DIKW structure \cite{kgsw}. Therefore, similar to data privacy protection \cite{social}, researchers began to pay attention to privacy protection of DIKW structure. The authors in \cite{dikwprivacy} uniformly categorize and extend the entities and relations in the multiple graphs of DIKW structure by explicit and implicit divisions of typed resources of data, information, and knowledge. This encryption in the multi-model graph of DIKW could efficiently improve the efficiency of privacy protection in DIKW structure as proved by the authors. However, until now not many efforts have been focused on the DIKW privacy protection \cite{mdik}.

Differential privacy is a popular approach in the data privacy protection \cite{dp}\cite{dpdeep}, aiming to deassociate the dataset with any individual' data. No matter if we add or remove any individual data from the dataset, the algorithm, including the training of deep learning model based on the dataset, does not change. The function of differential privacy $K$ is defined as Eq. \ref{eq:dp}:
\begin{equation}
\label{eq:dp}
		Pr[K(D_1) \in S] \leq exp(\varepsilon ) \times Pr[K(D_2)  \in S]
\end{equation}
where $D_1$ and $D_2$ differ on at most one element, and all $S \subseteq Range(K)$. \(K\) satisfies the demand of differential privacy as even if one individual data is leaked, the algorithm and the dataset do not change their characteristics. It is often used to inject noise, with Laplace distribution, and  controlled sensitivity into the original dataset to create an encrypted dataset for the purpose of differential privacy. At present, deep learning especially deep neural networks are trained by stochastic gradient descent for the same purpose of generating the encrypted dataset with better efficiency and effect \cite{dpsgd}. With the success of differential privacy in the data privacy protection, we would like to discuss the extended application of differential privacy in DIKW architecture.


 


\section{Purpose driven DIKW}
DIKW was originally proposed as the pyramid architecture \cite{dikw} with the data as the lowest model. However, in the recent research \cite{darc1}\cite{darc2}, researchers proposed that the different models of DIKW should be interactive, but the relations among models have not been clearly defined and discussed. In this paper, we propose novel relations among DIKW models: "purpose". The "purpose" is the organic power to link the different models of DIKW and unify them as a whole. We use the case to demonstrate the importance of the purpose. The $Data$, Computer, is considered as the tool in computation and the purpose is considered to do computing;by analogy the game console is the tool and the purpose is the entertainment or gaming. In the extreme case, we can use computer as the chopping board when the purpose is to cut food. Therefore, the purpose can significantly change the transmitted $Information$ by using the same $Data$. We use Figure \ref{fig:pdikw1} to illustrate the proposed principle: the purpose is formed by the internal relations among the models of DIKW. The architecture of the Purpose driven DIKW (PDIKW) is illustrated in Figure \ref{fig:pdikw2}.

\begin{figure}[ht]
\centering
\includegraphics[width=0.7\columnwidth]{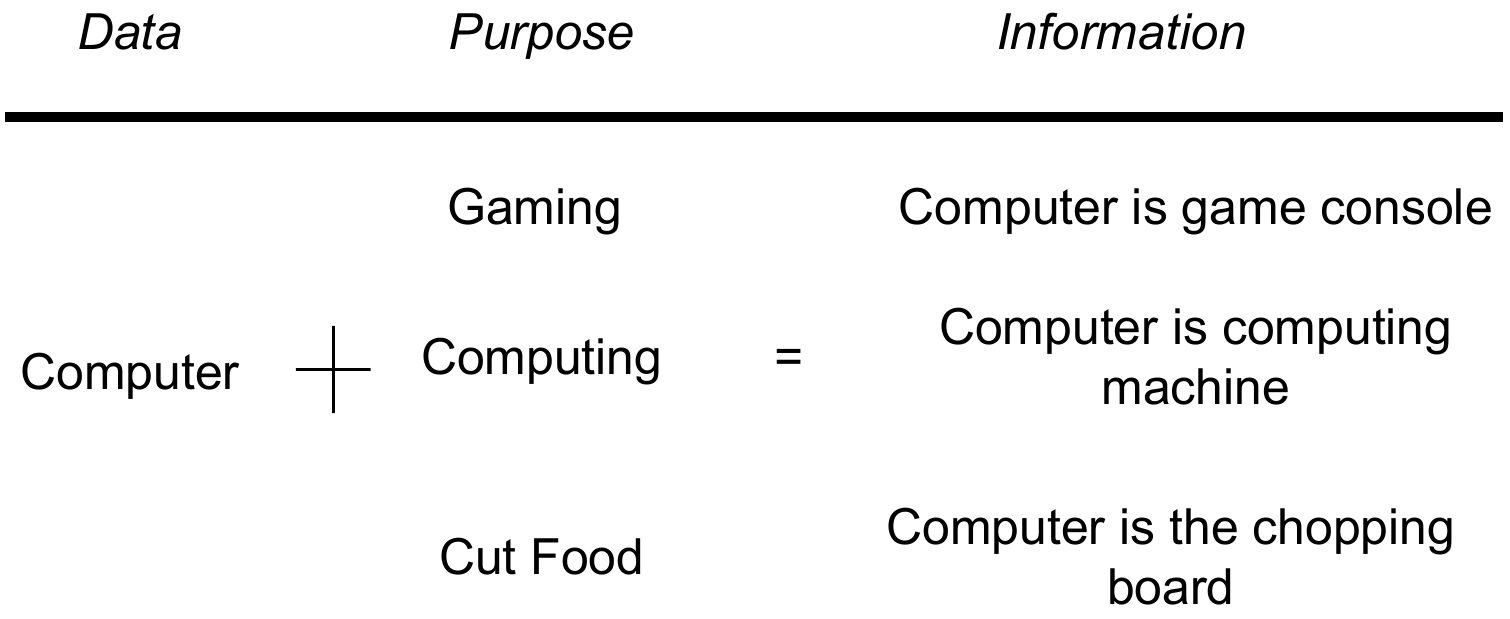}
\caption{PDIKW example}
\label{fig:pdikw1}
\end{figure}

\begin{figure}[ht]
\centering
\includegraphics[width=0.7\columnwidth]{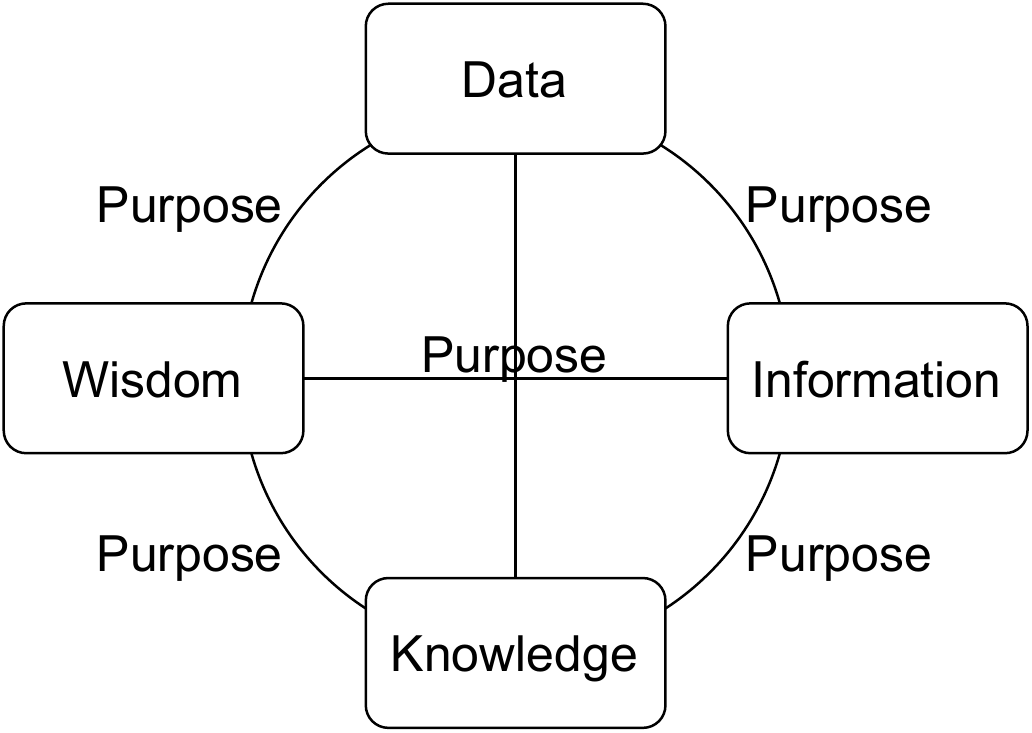}
\caption{PDIKW architecture}
\label{fig:pdikw2}
\end{figure}

\section{Differential Privacy for Purpose Driven DIKW}
In this section we would like to firstly review the related work of differential privacy to protect the data privacy, and then extend the philosophy of privacy protection by differential privacy to our DIKW model.

\subsection{Differential Privacy}
Privacy-preserving data analysis \cite{dp} is at crisis from a moral and regulation angle. So both the academics and industries almost never use data in their applications, with privacy from a specific group of people. Privacy-preserving data analysis, known as statistical disclosure control and private data analysis have the following drawbacks:

\begin{enumerate}
\item The data highly related to specific privacy is not suitable in statistical data analysis, including being used as training data for machine learning. Normally this kind of data is sensitive to specific data, which means that the modification, the increase and the removal of the specific data will influence the analysis and inference results from the data highly related to specific privacy. 
\item The dataset composed of privacy is not suitable to be used as the standard dataset. In the research and benchmark, we normally need the standard dataset as the ground truth. The privacy dataset is not good for this purpose as it is imbalanced.
\item The data with the privacy is not possible to be used for open source. The open source of the machine learning model and dataset is the trend in data science research and industry aty the moment. So the privacy data can be only used in the limited domains such as the company internal usage, which means it is hard to evaluate the data and the corresponding algorithm. 
\item The data with the privacy is facing the law and social challenge. With the concern to the privacy right, the society is paying more attention to their online privacy since the data leakage from the Internet is becoming increasingly serious. At the same time, the laws concerning the privacy protection such as the General Data Protection Regulation (EU GDPR)\cite{gdpr} and California Consumer Privacy Act (CCPA)\cite{ccpa} are made to prevent privacy abuse. The above laws increase bring challenges to the analysis of privacy-preserving data.
\end{enumerate}

Therefore, the researchers are using different approaches to mask and remove the privay data and sensitive data from the datasets \cite{datamasking}. We would review several important approaches:
\begin{enumerate}
\item Data substitution: The data that is not related to the computing can sometimes replaced for example the name.
\item Data encryption: The password can be encrypted without influencing the other data. The encryption would change the look and feel of the data, which also needs more computing time and resources for encryption and decryption.
\item Data shuffling: The data shuffling could remove the data ordering together with the temporal information among the data items. While the data shuffling would not work if the dataset items are fewer, for example a dataset only has 10 records.
\item Data noising: Gaussian noise is normally equally added to each value of the dataset in order to remove the uniqueness of each value in the dataset. Differential privacy has been proved to be a successful approach for this category.
\end{enumerate}

We have introduced the basic definition for differential privacy in Eq. \ref{eq:dp}. The demand of a successful algorithm of differential privacy comes from the fact that the noise can be removed by the many responses of the algorithm to the datasets. (To be decided if reviewing approaches of DP). With the success of deep learning, the approach nehing it has been used for in differential privacy applications. Abadi et al.\cite{dpdeep} introduced a new approach with a more efficient cost of computing and a tighter privacy loss, working with the machine learning framework TensorFlow. This  approach is based on Differentially Private SGD Algorithm as described in Fig. \ref{fig:dpsgd} \cite{dpdeep}.

\begin{figure}[ht]
\centering
\includegraphics[width=0.9\columnwidth]{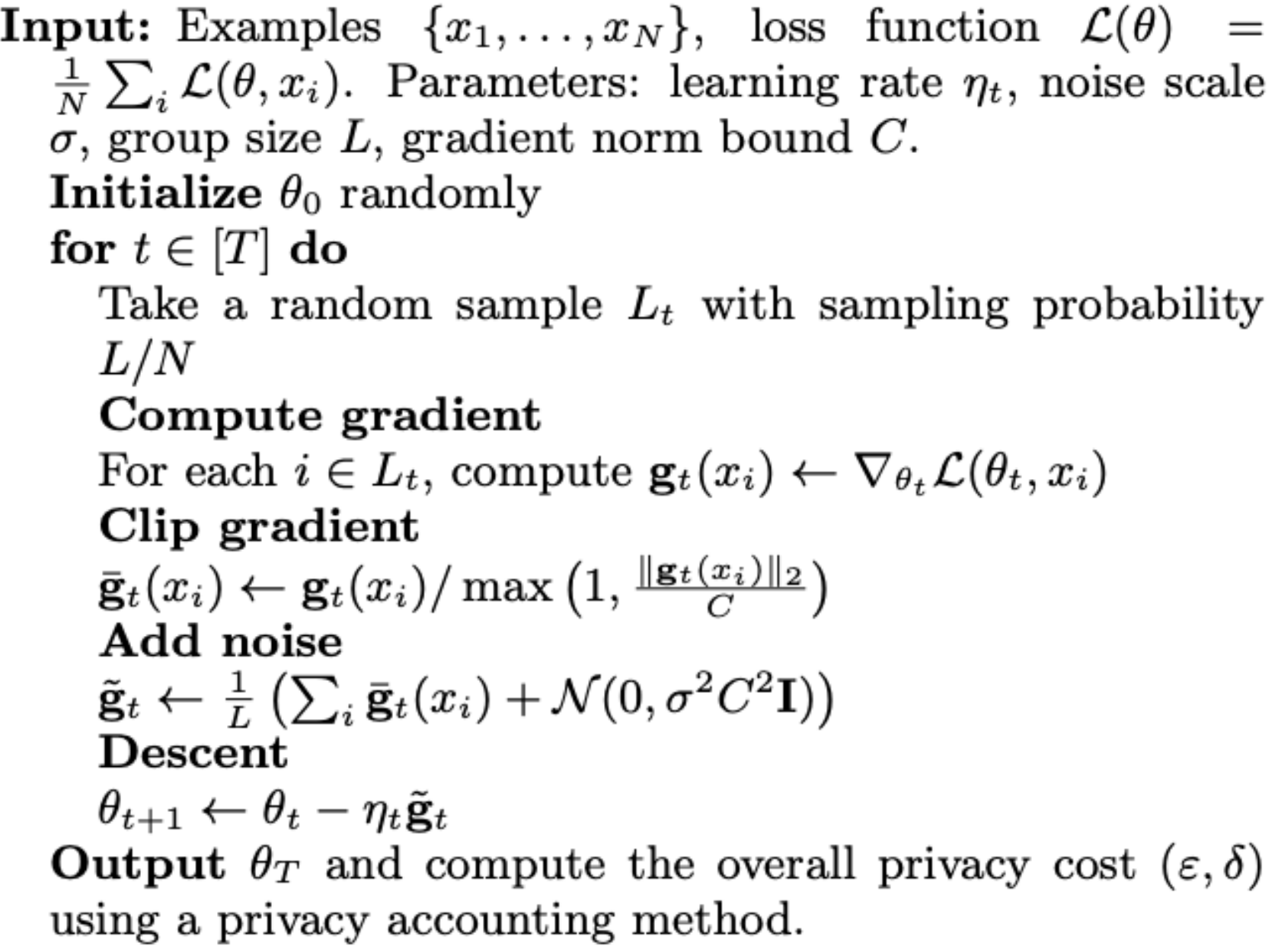}
\caption{Differentially Private SGD}
\label{fig:dpsgd}
\end{figure}

\subsection{DIKW Differential Privacy}

The example of DIKW framework \cite{dikwprivacy} is described in Fig. \ref{fig:DIKWframework}. This framework originates from the real world extending the existence of things.  The human semantics as the relation connect the entities. DIKW could represent 5 Ws: Who, What, When, Where, and Why, and connect to the data, information and knowledge in DIKW.

\begin{figure*}[ht]
\centering
\includegraphics[width=0.9\textwidth]{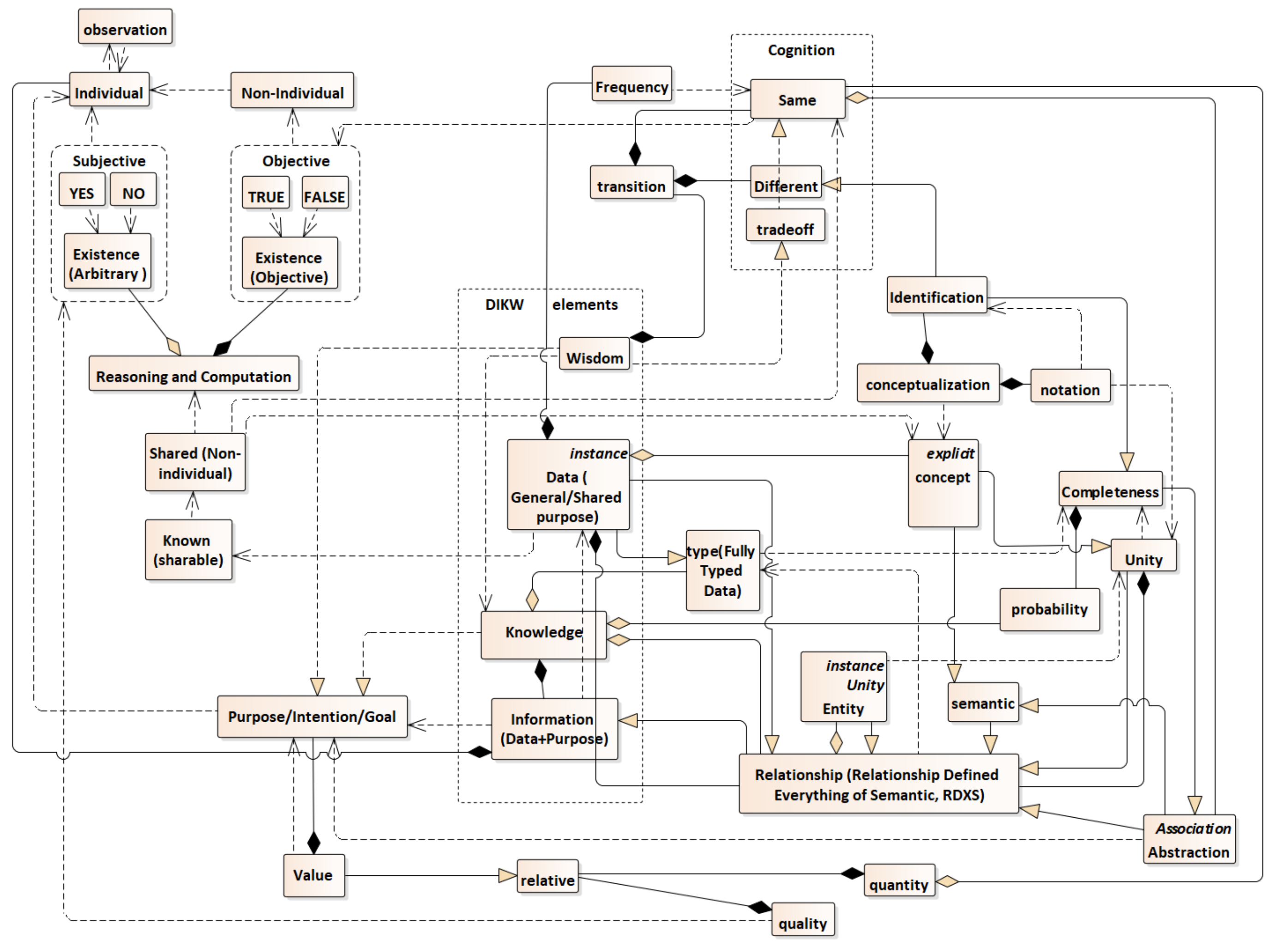}
\caption{DIKW framework}
\label{fig:DIKWframework}
\end{figure*}

Differential privacy is normally applied to the data, especially to the training data in deep learning approaches, when we consider the data, information and knowledge models in the DIKW architecture. Differential privacy could be applied to each of these three models separately and jointly:
\begin{enumerate}
\item Data Differential Privacy (DDP). Since the data in DIKW is highly close to the $Who, When$ and $Where$ in 5$W$s, we could only apply the differential privacy to the value items closely related to $Who, When$ and $Where$, in order to save the computing cost. The principle is formalized into Eq. \ref{eq:ddp}.
\[ D_{DDP} =
          \begin{cases}
            dp(D) &\text{if } D\in [Who,When,Where] \\
            D &\text{if } D \notin [Who,When,Where]
          \end{cases}\label{eq:ddp}
        \]
where $dp$ represents the processing of differential privacy to the item D in the data of DIKW.
\item Information Differential Privacy (IDP). Similarly, the information in DIKW represents the meaning of $What$, so we could apply the differential privacy to the items representing What in 5Ws in the information. This is represented by Eq. \ref{eq:idp}.
    \[ I_{IDP} =
          \begin{cases}
            dp(I) &\text{if } I\in [What] \\
            I &\text{if } I \notin [What]
          \end{cases}\label{eq:idp}
        \]
where $dp$ represents the processing of differential privacy to the item I in the information of DIKW.
\item Knowledge Differential Privacy (KDP). We define the transformed knowledge by differential privacy in Eq. \ref{eq:kdp}.
    \[ K_{KDP} =
          \begin{cases}
            dp(K) &\text{if } K\in [How] \\
            K &\text{if } K \notin [How]
          \end{cases}\label{eq:kdp}
        \]
Correspondingly, we can only apply differential privacy to the How related item in the $Knowledge$ of DIKW.
\item Data-Information Differential Privacy (DIDP). DIDP is the joint differential privacy application on both Data and Information in DIKW to all items related to $Who, When, Where$ and $What$. 

    $t_{DIDP}$ = 
    \[ 
          \begin{cases}
            dp(t)&\text{if } t\in [Who,When,Where,What] \\
            t&\text{if } t \notin [Who,When,Where,What]
          \end{cases}\label{eq:didp}
        \]
\item Information-Knowledge Differential Privacy (IKDP). IKDP is the joint differential privacy application on both $Knowledge$ and $Information$ in DIKW to all items related to $How$ and $What$. 

    $t_{IKDP}$ = 
    \[ 
          \begin{cases}
            dp(t)&\text{if } t\in [How,What] \\
            t&\text{if } t \notin [How,What]
          \end{cases}\label{eq:ikdp}
        \]
\item Data-Information-Knowledge Differential Privacy (DIKDP). DIKDP is the mode with the heavy burden of computing cost since it needs to apply differential privacy to all the items in DIKW.

    $t_{DIKDP}$ = 
    \[ 
          \begin{cases}
            dp(t)&\text{if } t\in [5Ws] \\
            t&\text{if } t \notin [5Ws]
          \end{cases}\label{eq:dikdp}
        \]
\item Purpose Differential Privacy (PDP). PDP disasscoiates the models relations by the purpose in Purpose driven DIKW in order to protect the privacy.

    $t_{PDP}$ = 
    \[ 
          \begin{cases}
            dp(p)&\text{if } p\in P \\
            p&\text{if } p \notin P
          \end{cases}\label{eq:pdp}
        \]

\end{enumerate}

From DIKW view and saving computing resources perspective, we propose to apply the IDP firstly if it satisfies the demand of privacy protection since the value item related to $What$ in the Information is the least, while KDP related to the knowledge is too high level and still discloses too much data; DDP will take more computing time and resources compared to IDP and KDP. DIDP, IKDP. DIKDP with the joint differential privacy application will take more computing time, so they should be applied only in necessary situations. More practical discussion will be conducted in the section of case study.

\section{Spatial-Temporal Swarm Differential Privay for DIKW }
In this section we will firstly review the history and application of Swarm Intelligence (SI) \cite{swarm}, an integral part of Artificial Intelligence (AI), especially the Particle Swarm Optimization (PSO)\cite{pso} in Swarm Intelligence. Then we propose the architecture to integrate swarm intelligence with differential privacy for DIKW in the previous section in order to achieve more effective and efficient privacy protection. Finally we propose to add the spatial-temporal information consideration into the proposed architecture.
\subsection{Swarm Intelligence}
Swarm intelligence is the bio-inspired computing algorithm by mimicking the behavior of a cluster of animals or insects. Gerardo Beni and Jing Wang initially proposed swarm intelligence when they researched on the cellular robotic system, embracing the algorithm characteristics of flexibility and versatility. When a cluster of animals or insects are together for living and moving, each animal or insect could adapt itself to the behaviour of the whole cluster. Swarm intelligence originated from and borrowed this feature. In the past years swarm intelligence has been widely applied in business planning, computer science, industrial applications etc.

Inside the algorithms of swarm intelligence, ant colony optimization, fish swarm optimization, particle swarm optimization, bee-inspired algorithms, bacterial foraging optimization and firefly algorithms are popular categories \cite{swarm2}. Swarm intelligence included both stationary optimization and dynamic optimzation until now. Ant colony optimization is for discrete optimization, while the other above listed algorithms are for continuous optimization.

We will briefly overview several popular algorithms of swarm optimization. Ant colony optimization mimick the behavior of ant group finding the nearest route from the nest to the food. Each ant has an optimized route to send to background while the cluster would consider all the potential solutions and decide one optimized solution for the whole cluster. An example of ant colony optimization is shown in Fig. \ref{fig:ant}. Artificial bee colony is different with ant colony optimization at the point that the food source could be of multiple choices and varies according to time and selection. In fish swarm optimization each artificial fish can view its neighbors' behavior when each artificial fish is looking for the best food so its motion is influenced by the local neighbors. We show the illustration of fish swarm optimization in Figure \ref{fig:fish}. In firefly algorithms the view range of one firefly to the neighbour fireflies is influenced by the landscape of the problem while the firefly attracts the motions of each other.

\begin{figure}[ht]
\centering
\includegraphics[width=0.7\columnwidth]{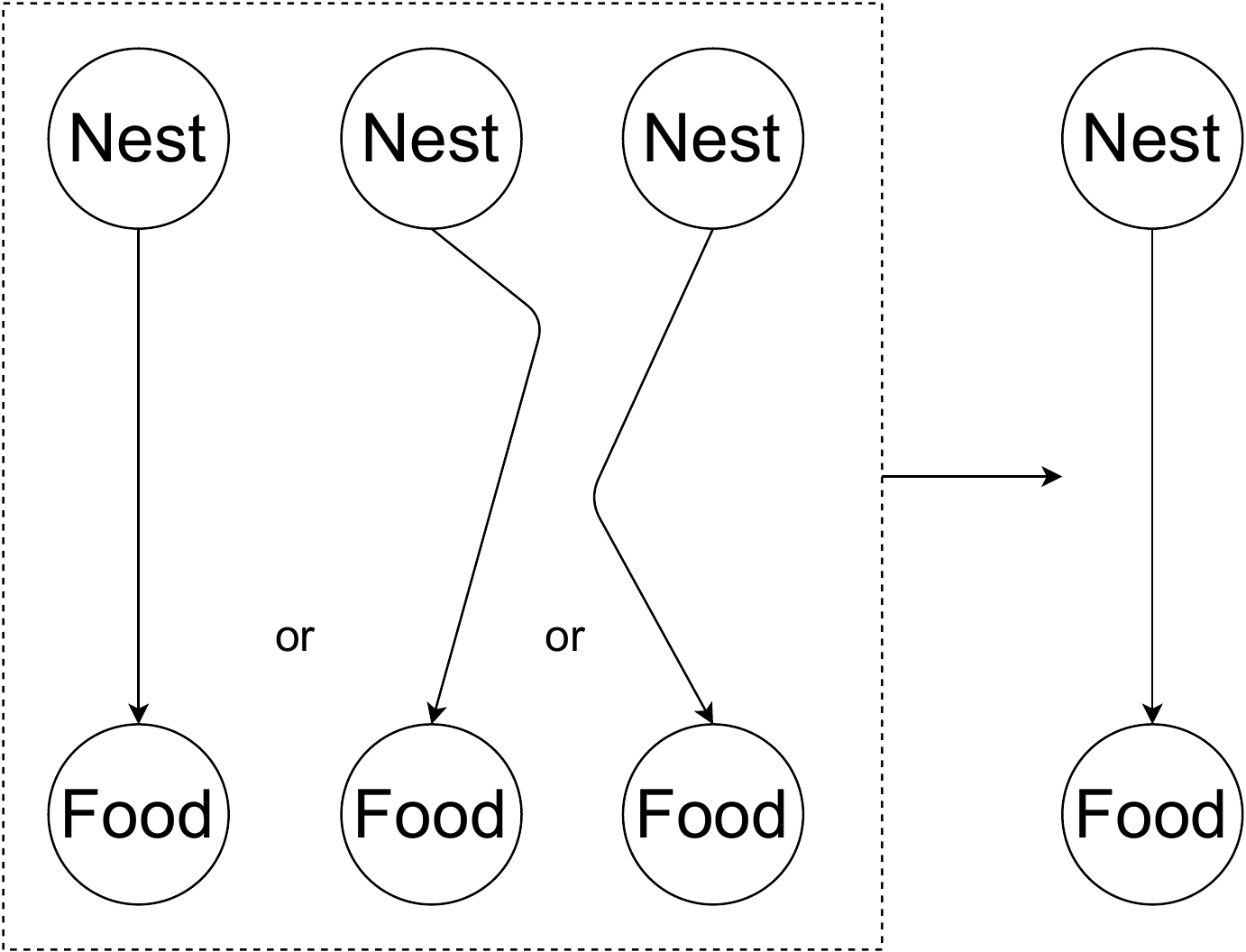}
\caption{Ant colony optimization}
\label{fig:ant}
\end{figure}

\begin{figure}[ht]
\centering
\includegraphics[width=0.5\columnwidth]{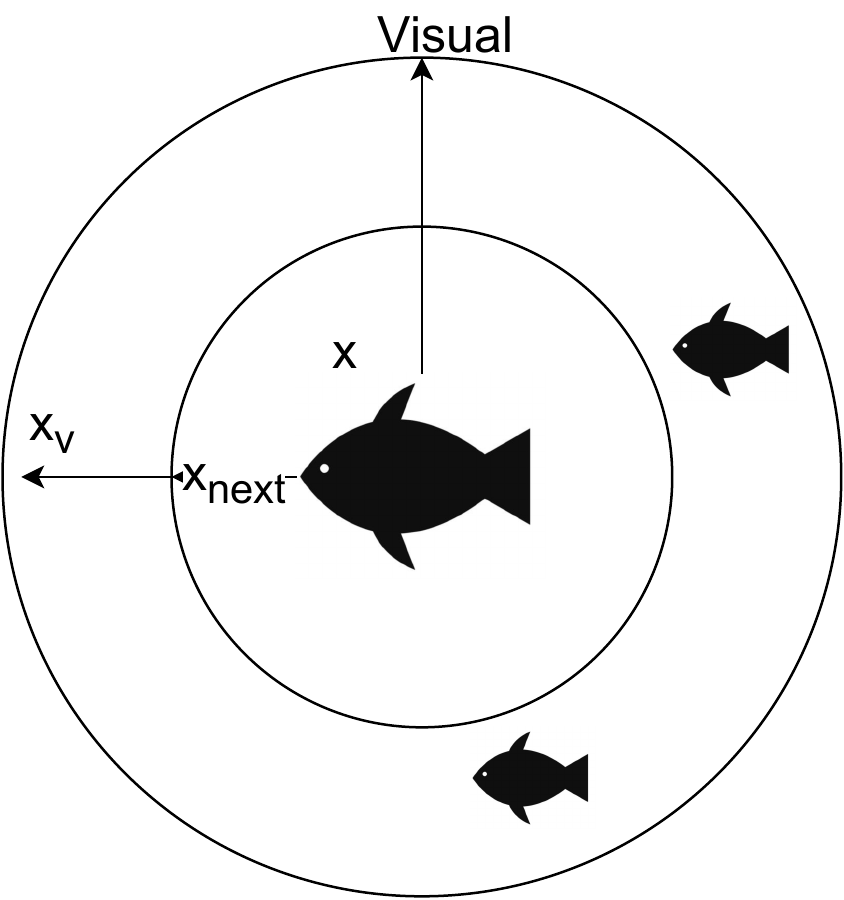}
\caption{Fish swarm optimization}
\label{fig:fish}
\end{figure}

Particle Swarm Optimization (PSO) was initially proposed in 1995\cite{psoo} for the continuous optimization problems. Similar to the ant colony optimization, each particle is a potential solution to the problem, with the velocity and position vectors, that are optimized at each step of the progess according to the positions of the particles and the swarm. PSO has two kinds: the local best model optimizing considering the local neighbors and the global best model optimizing according to the whole swarm. So, in PSO \cite{pso} the particle decides the next step in its motion with both local best and global best: the own best position and the global best or their neighborhood's best. The particle in each step of the progress has its current position $x_i(t)$ and the velocity $v_i(t)$, and the following position is defined by Eq. \ref{eq:pso}.
    \begin{equation}
    \label{eq:pso} 
        x_i(t+1) = x_i(t)+v_i(t+1)
    \end{equation}
where $v_i(t) = v_i(t-1)+c_1r_1(localbest(t)-x_i(t-1))+c_2r_2(globalbest(t)-x_i(t-1))$. Acceleration coefficients are $c_1$ and $c_2$, and random vectors are $r_1$ and $r_2$.The structure of PSO algorithm is illustrated in Figure \ref{fig:psof}  \cite{psog}.

\begin{figure}[ht]
\centering
\includegraphics[width=0.3\columnwidth]{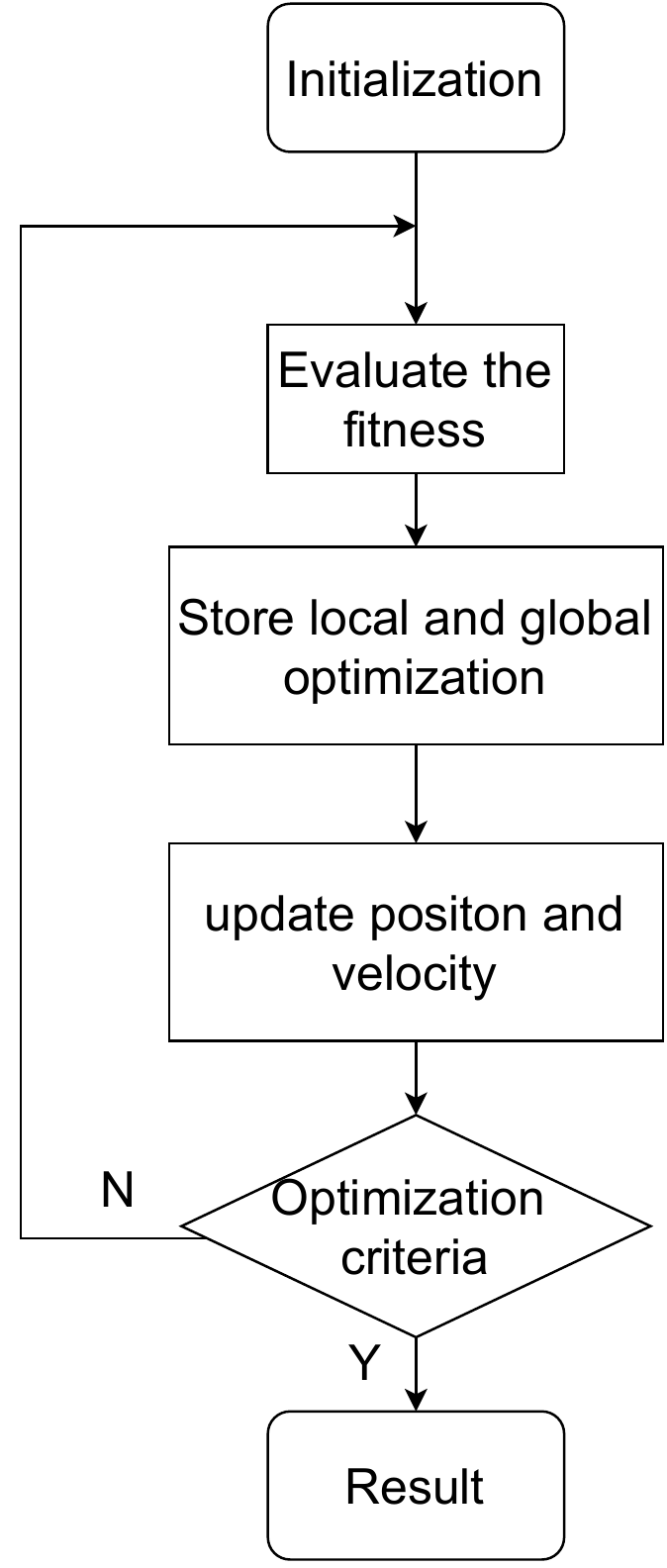}
\caption{Particle Swarm Optimization}
\label{fig:psof}
\end{figure}

\subsection{Swarm Differential Privay for DIKW}
Particle Swarm Optimization (PSO) considers the optimization of both local best and global best as described in the previous section, which considers its own status, the neighboring particles and all the particles for each particle. PSO has been widely used in multiple applications \cite{psoh}, and is the proved successful optimization algorithm.

We propose to apply PSO in the differential privacy for DIKW. The differential privacy to all the items of the selected models or all models in DIKW - as described in DDP, IDP, KDP, DIDP, IKDP and DIKDP means the computing workload, computing time and the influence to the efficiency of differential privacy. Therefore, we propose the novel application of PSO to the differential privacy for DIKW architecture in the decision of differential privacy to each item in DIKW, which means that only selected items by PSO considering the local optimization and global optimization will be "masked" by differential privacy. We use Figure \ref{fig:dpdikw1} to illustrate the proposed principle, and the solid dot means the application of differential privacy to that item in DIKW. We regard the items in DIKW shown in Figure \ref{fig:dpdikw1} as the particles in PSO, while all the items in DIKW as the swarm. The position of each particle $x_i(t)$ is regarded as the boolean status of differential application, while the velocity $v_i(t)$ is the transformation of boolean status. The Eq.2 of PSO is adapted and changed to Eq. \ref{eq:psodikw}.

    \begin{equation}
    \label{eq:psodikw} 
        x_i(t+1) = x_i(t)*v_i(t+1)
    \end{equation}
where $v_i \in \{-1,1\}$ and $x_i \in \{-1,1\}$. $x_i=1$ means the application of differential privacy while $x_i=-1$ means the non-application of different privacy to the corresponding item in DIKW. $v_i=1$ means not transforming the status of differential privacy application, while $v_i=-1$ means the transformation of differential privacy application to the item in DIKW. And $v_i(t)=v_i'(t)*x_i(t-1)$, and $v_i'(t)$ is defined in Eq. \ref{eq:velo}:

    \begin{equation}
        v_i'(t) =           
        \begin{cases}
            -1 &\text{if } v_i''(t)<0 \\
            1 &\text{if } v_i''(t)>0
          \end{cases}
    \label{eq:velo}
    \end{equation}
where $v_i''(t)=v_i(t-1)+c_1r_1(localbest(t)-x_i(t-1))+c_2r_2(globalbest(t)-x_i(t-1))$.

By the progressive optimization of PSO to differnet privacy for DIKW, we could identify the most important items in DIKW. Thus we could save the time of differential privacy while preserve the effect and accuracy of differential privacy. The converge condition of PSO is defined as the serious increase of data variance when the number of valid times with differential privacy shrinks.

\begin{figure}[ht]
\centering
\includegraphics[width=0.7\columnwidth]{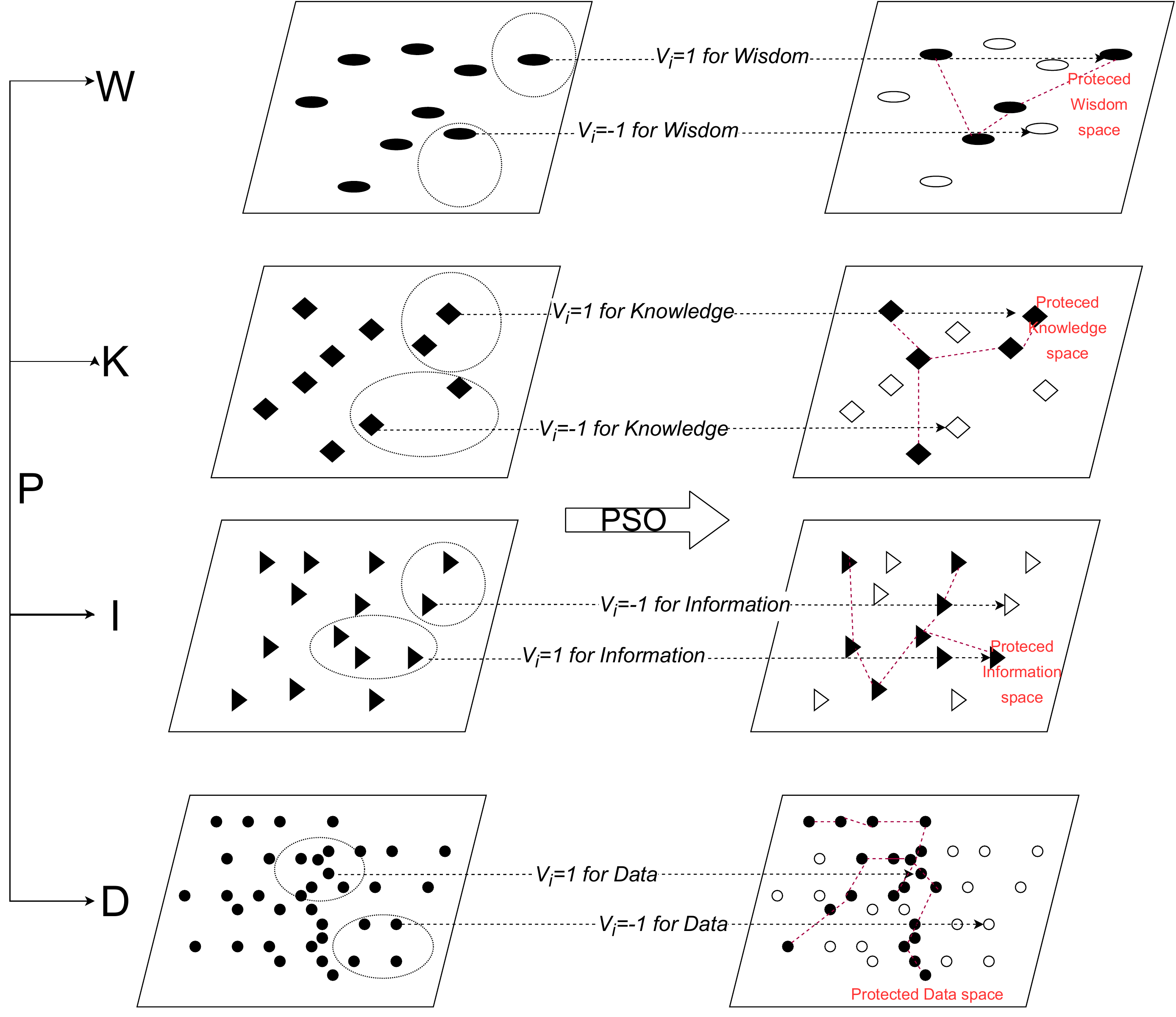}
\caption{PSO optimized differential privacy for DIKW}
\label{fig:dpdikw1}
\end{figure}

We can also use the final valid items with differential privacy as the feature to discover the relations among the models of DIKW in order to decide the model in differential privacy. For this purpose, we should begin from the bottom model $Data$. If the remaing valid items in $Data$ has strong semantic relation to the upper model Information, both $Data$ and $Information$ model should be used in the differential privacy, which means DIDP should be used in the differential privacy for DIKW. In order to decide the model among DDP, IDP, KDP, DIDP, IKDP and DIKDP, we could launch the process from $Data$ model, $Information$ model and $Knowledge$ model. The associated models would be used to decide the model of differential privacy for DIKW. PDP is unique and different with other proposed approaches. The principle is shown in Figure \ref{fig:dpdikw2}.

\begin{figure}[ht]
\centering
\includegraphics[width=0.8\columnwidth]{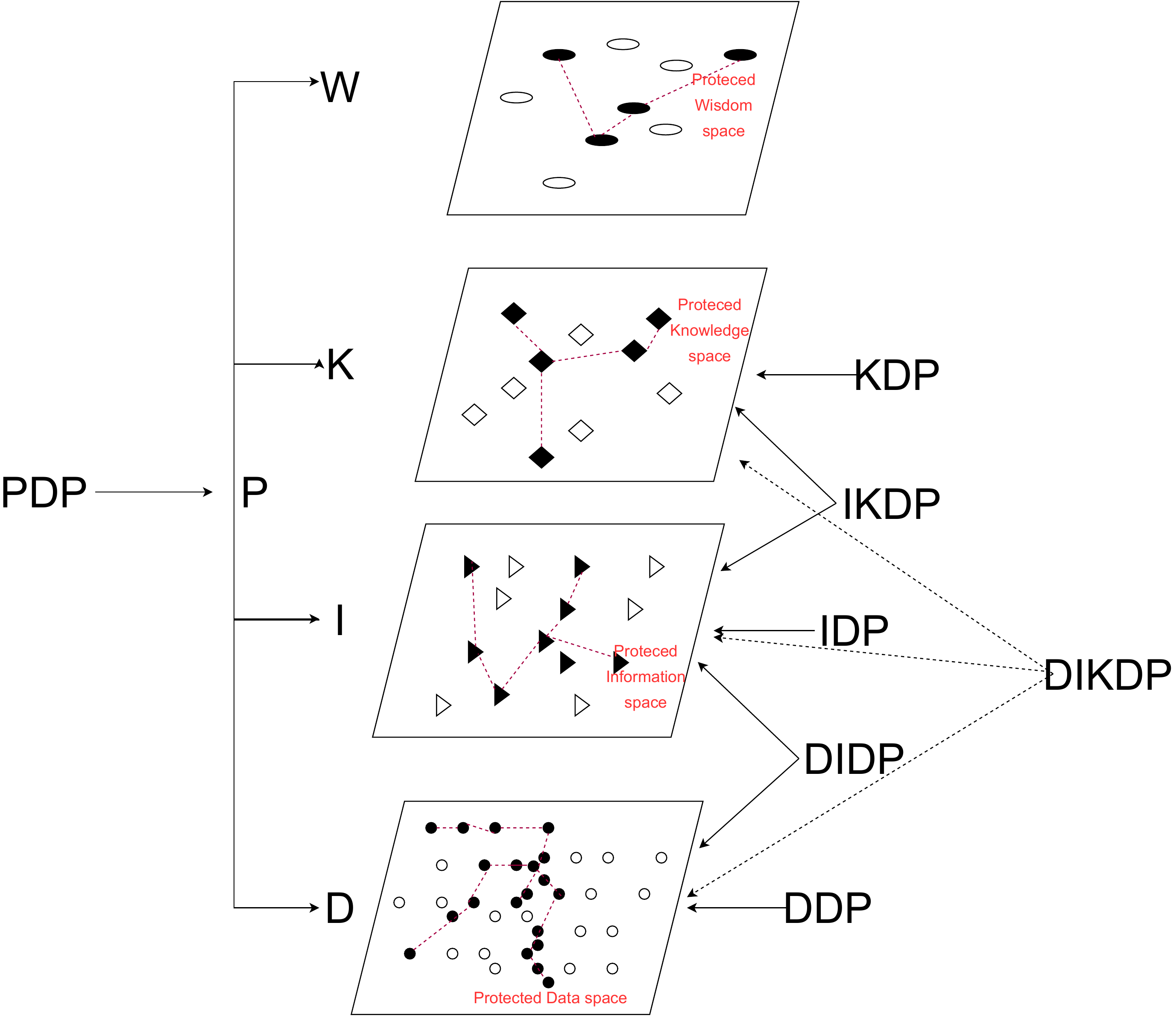}
\caption{PSO modelling in differntial privacy for DIKW}
\label{fig:dpdikw2}
\end{figure}

\subsection{Spatial-Temporal Swarm Differential Privay for DIKW}
The spatial and temporal relations among the items in different models of DIKW are important information to protect privacy. Here we use the example to discuss. We assume the we have the DIKW information from the social media. The spatially neighboring users normally have some kind of the same trends that should be especially processed by differential privacy. The recent information from social media definitely represents the user's true preferences than the information recorded historically. 
We could apply the spatial-temporal information into PSO Differential privacy for DIKW similary in two ways. Firstly, we could apply spatial-temporal information in the model decision of PSO modelling in differential privacy for DIKW as shown in Figure \ref{fig:dpdikw3}. Since the model decision is from the semantic relations among models in DIKW, the spatial-temporal information could highly influence these semantic relations. The spatial-temporal information here includes but not it is not limited to the data time, the physical position, the text similarity and the data recording time in the database. Secondly, we could optimize the equation in swarm differential privacy for DIKW, Eq. \ref{eq:psodikw} by the spatial-temporal information and adapte Eq. \ref{eq:velo} to Eq. \ref{eq:veloe}.
    \begin{equation}
        v_i'(t) =           
        \begin{cases}
            -1 &\text{if } v_i''(t)<0 \\
            1 &\text{if } v_i''(t)>0
          \end{cases}
    \label{eq:veloe}
    \end{equation}
where $v_i''(t)=v_i(t-1)+c_1r_1(localbest'(t)-x_i(t-1))+c_2r_2(globalbest'(t)-x_i(t-1))$. Both $localbest'$ and $globalbest'$ take into account of the spatial-temporal information in the measurement of the similarity.

The spatial-temporal semantics improve the performance the differential privacy for DIKW in the aspect of the data semantics in each model and among models of DIKW. It helped  accurately define the item relations among DIKW models and save the computing load of differential privacy for DIKW from the macro way. While in the micro-way the spatial-temporal semantics participate into the PSO computation.

\begin{figure}[ht]
\centering
\includegraphics[width=0.4\columnwidth]{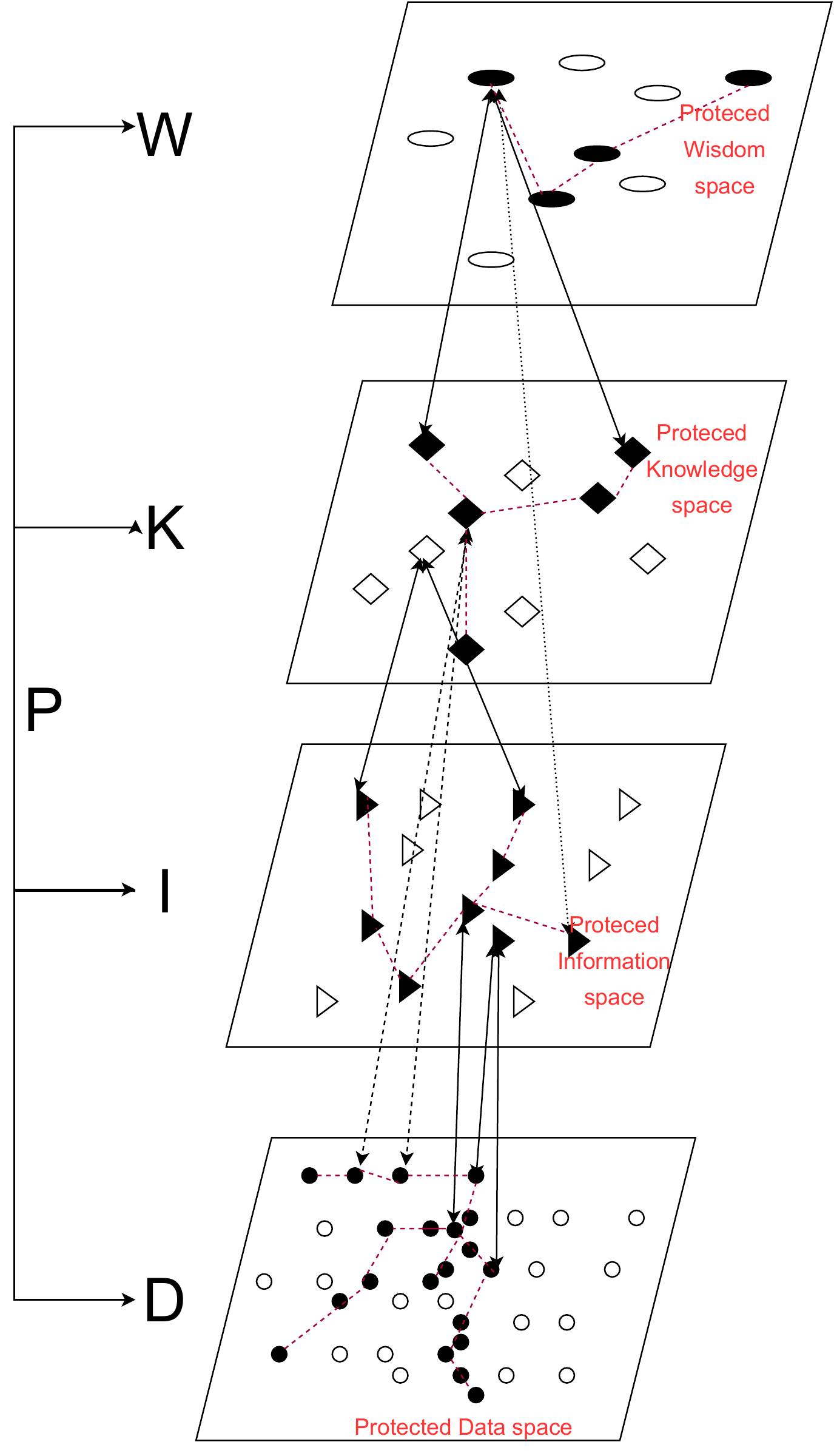}
\caption{The spatial-temporal semantics among models in DIKW}
\label{fig:dpdikw3}
\end{figure}

\section{Application and Case Study}

In the proposed approach, Particle Swarm Optimization could efficiently reduce the DIKW items with differential privacy application according to the defined criteria, in the semantically prefered models of DIKW architecture. The proposed approach could improve the data privacy efficiency but at the same time broadly keep the effectiveness. According to our knowledge, the proposed approach originally combined the advantages of swarm intelligence with differential privacy for the DIKW information and data privacy protection. We know by heart that the proposed architecture is a little complex for some variants, so in the future it will be a part of our work to simplify the proposed approach and condense into the best and optimized approach.

In order to prove the proposed approach, we use the extended dataset based on Iris dataset \cite{iris}. We artificially enhance the Iris dataset to DIKW architecture for the proposed approach. Since the differential privacy by deep learning has been conducted for years by the researchers, we have many variants of differential privacy algorithms and libraries to exploit. The experiment is conducted using the differential privacy libarary from IBM \cite{dpl}. We show the accuracy of the remaining items of different percentages after PSO application in Figure \ref{fig:exp}. We can see that with the increase of processing epsilons, the differential privacy to the partial items in DIKW could still effectively keep the accuracy of differential privacy. Since the open-source DIKW dataset is normally missing in the research community, it is hard to compare our proposed approach with other algorithms. We will build an open-source dataset especially used for DIKW privacy protection as the target in the future research. Thus, we could use this standard benchmark to evaluate different algorithms for DIKW. 

\begin{figure}[ht]
\centering
\includegraphics[width=1.0\columnwidth]{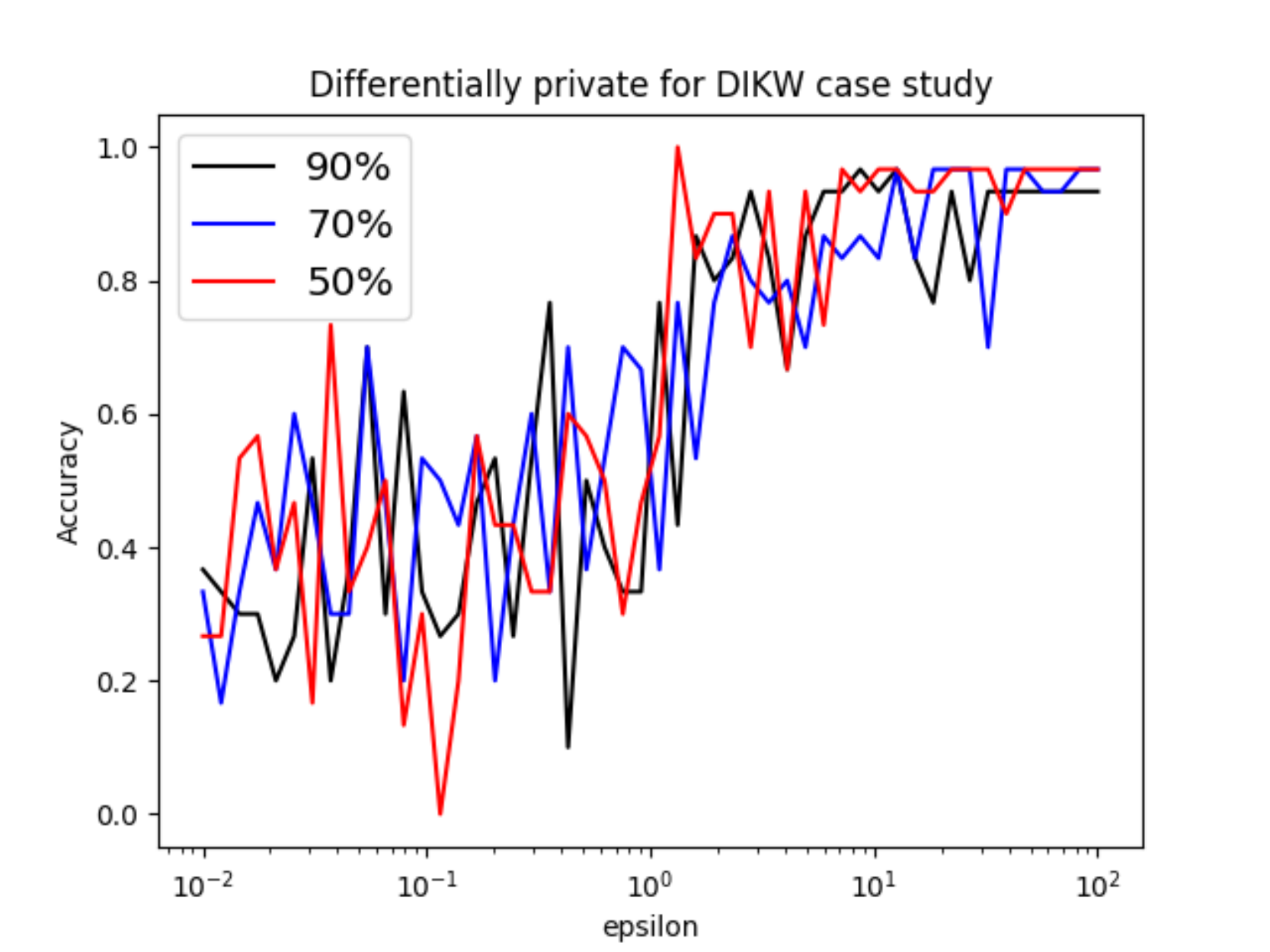}
\caption{The case study}
\label{fig:exp}
\end{figure}

\section{Conclusion}
Privacy protection has attracted many attentions from edge computing to mobile computing, with most efforts put to the data privacy protection. We have not seen much research on the privacy protection on multiple modals DIKW model. Conceptual Modal based DIKW protection is naturally more complex than the current popular data protection since data modal is only one modal of DIKW architecture. In our DIKW mddel we could exploit more semantic relations among the modals of DIKW and inside each modal of DIKW. In this paper, we originally propose the crossing DIKW modals' privacy protection by extending differential privacy. Furthermore, we use Particle Swarm Optimization to enhance the efficiency of differential privacy from two aspects: multiple modals of DIKW model and PSO optimization progress, which optimizes the differential privacy from macro aspect to micro aspect, and from the global best to the local best potentially crossing multiple dimensions and scales even meso-scales .

Aiming at cognitive integrity and integration overload for complex content identification, modeling, processing, and service optimization in the context of massive content interaction in multi-dimensional, multi-modal, multi-scale physical and digital space, especially towards covering the digital twins landscape, we have proposed the architecture of fusing the DIKW modals, differential privacy and Swarm Intelligence for DIKW privacy protection. Actually following the proposed framework, more research could be conducted in the future to exploit the effective and efficient crossing modal DIKW privacy protection. The semantic relations among different DIKW models could be studied by different novel approaches like BERT \cite{bert}. We use PSO in Swarm Intelligence in this paper, but we may use different swarm intelligence algorithms like fish swarm algorithms to conduct the research and experiments. We could similarly change differential privacy to other algorithms of data masking and encryption. Therefore in this paper we wish to propose the framework of DIKW privacy protection other than the specific algorithm. We will continue the study of other algorithms combination under the proposed framework in the future.


\ifCLASSOPTIONcompsoc
  \section*{Acknowledgments}
\else
\fi

Supported by Natural Science Foundation of China Project (No. 61662021 and No.72062015), Hainan Provincial Natural Science Foundation Project No.620RC561, Hainan Education Department Project (No. Hnjg2021ZD-3 and No. Hnky2019-13) and Hainan University Educational Reform Research Project (No. HDJY2102, No. HDJWJG03)



%

\end{document}